\pdfoutput=1

\documentclass[11pt]{article}

\usepackage[]{acl}

\usepackage{times}
\usepackage{latexsym}

\usepackage[T1]{fontenc}

\usepackage[utf8]{inputenc}
\usepackage{adjustbox}

\usepackage{microtype}

\usepackage{graphicx}
\usepackage{booktabs}
\usepackage{multirow}
\usepackage{enumitem}
\usepackage{makecell}
\usepackage[ruled,vlined,linesnumbered]{algorithm2e}
\SetAlCapNameFnt{\small}
\SetAlCapFnt{\small}
\usepackage[noend]{algpseudocode}
\makeatletter
\def\ALG@special@indent{%
    \ifdim\ALG@thistlm=0pt\relax
        \hskip-\leftmargin
    \else
        \hskip\ALG@thistlm
    \fi
}
\newcommand{\Input}[1]{\item[]\noindent\ALG@special@indent \textbf{Input}\ #1}
\renewcommand{\While}[1]{\item[]\noindent\ALG@special@indent \textbf{while}\ #1 \textbf{do}}
\newcommand{\Statement}[1]{\item[]\noindent\ALG@special@indent #1}
\newcommand{\EndKernel}{\item[]\noindent\ALG@special@indent \textbf{End Kernel}}
\makeatother

\usepackage{amsmath}
\DeclareMathOperator*{\argmax}{arg\,max}

\usepackage{amsfonts}
\newcommand{\mb}{\mathbf}

\newcommand{\cutsectionabove}{}
\newcommand{\cutsectionbelow}{}
\newcommand{\cutsubsectionabove}{}
\newcommand{\cutsubsectionbelow}{}
\newcommand{\cutfigurebelow}{}
\newcommand{\procel}{ProceL}
\newcommand{\crosstask}{CrossTask}
\newcommand{\subtask}[1]{\texttt{#1}}
\newcommand{\script}{task graph}
\newcommand{\Script}{Task Graph}
\newcommand{\scripts}{task graphs}
\newcommand{\andorgraph}[1]{{\footnotesize\textsf{#1}}}
\usepackage{array}
\newcolumntype{L}[1]{>{\raggedright\let\newline\\\arraybackslash\hspace{0pt}}m{#1}}
\newcolumntype{C}[1]{>{\centering\let\newline\\\arraybackslash\hspace{0pt}}m{#1}}
\newcolumntype{R}[1]{>{\raggedleft\let\newline\\\arraybackslash\hspace{0pt}}m{#1}}

\DeclareRobustCommand\onedot{\futurelet\@let@token\@onedot}
\def\onedot{.}
\def\eg{\emph{e.g}\onedot} 
\def\ie{\emph{i.e}\onedot}

\def\pcond{G}

\usepackage{cleveref}
\usepackage{pgfplots} %
\pgfplotsset{width=6.5cm, compat=1.6}

\newcommand{\lajan}[1]{}

\usepackage{caption}
\usepackage{subcaption}

\usepackage{microtype}

\title{\fontsize{14}{14}\selectfont Unsupervised Task Graph Generation from Instructional Video Transcripts}

\author{Lajanugen Logeswaran$^1$, Sungryull Sohn$^1$, Yunseok Jang$^{1,2}$\thanks{$\,\,\,$Work done during an internship at LG AI Research}\hspace{0.3em}, Moontae Lee$^1$, Honglak Lee$^1$ \\[0.5em]
$^1$LG AI Research \hspace{1em} $^2$University of Michigan, Ann Arbor
}

\begin{document}
\maketitle

\begin{abstract}
This work explores the problem of generating \emph{\scripts}~of real-world activities. %
Different from prior formulations, we consider a setting where text transcripts of instructional videos performing a real-world activity (e.g., making coffee) are provided and the goal is to identify the key steps relevant to the task as well as the dependency relationship between these key steps.
We propose a novel \script~generation approach that combines the reasoning capabilities of instruction-tuned language models along with clustering and ranking components to generate accurate \scripts~in a completely unsupervised manner.
We show that the proposed approach generates more accurate \scripts~compared to a supervised learning approach on tasks from the \procel~and \crosstask~datasets.

\end{abstract}

\vspace{0.2em}
\cutsectionabove
\section{Introduction}
\cutsectionbelow

Tasks in the real-world are composed of multiple key steps with specific dependencies that dictate the order in which they can be performed (\eg, one has to \emph{check for breathing} before \emph{performing CPR}).
Exposing these dependencies between key steps has many downstream applications including assisting human users in troubleshooting and building artificial agents that efficiently learn and perform new tasks.
However, information about tasks is typically available in unstructured and noisy form in the wild (\eg, `how to' descriptions or instructional video transcripts), presenting a major challenge in extracting structured representations. %

\begin{figure}[t!]
    \centering
        \centering
        \includegraphics[width=0.48\textwidth]{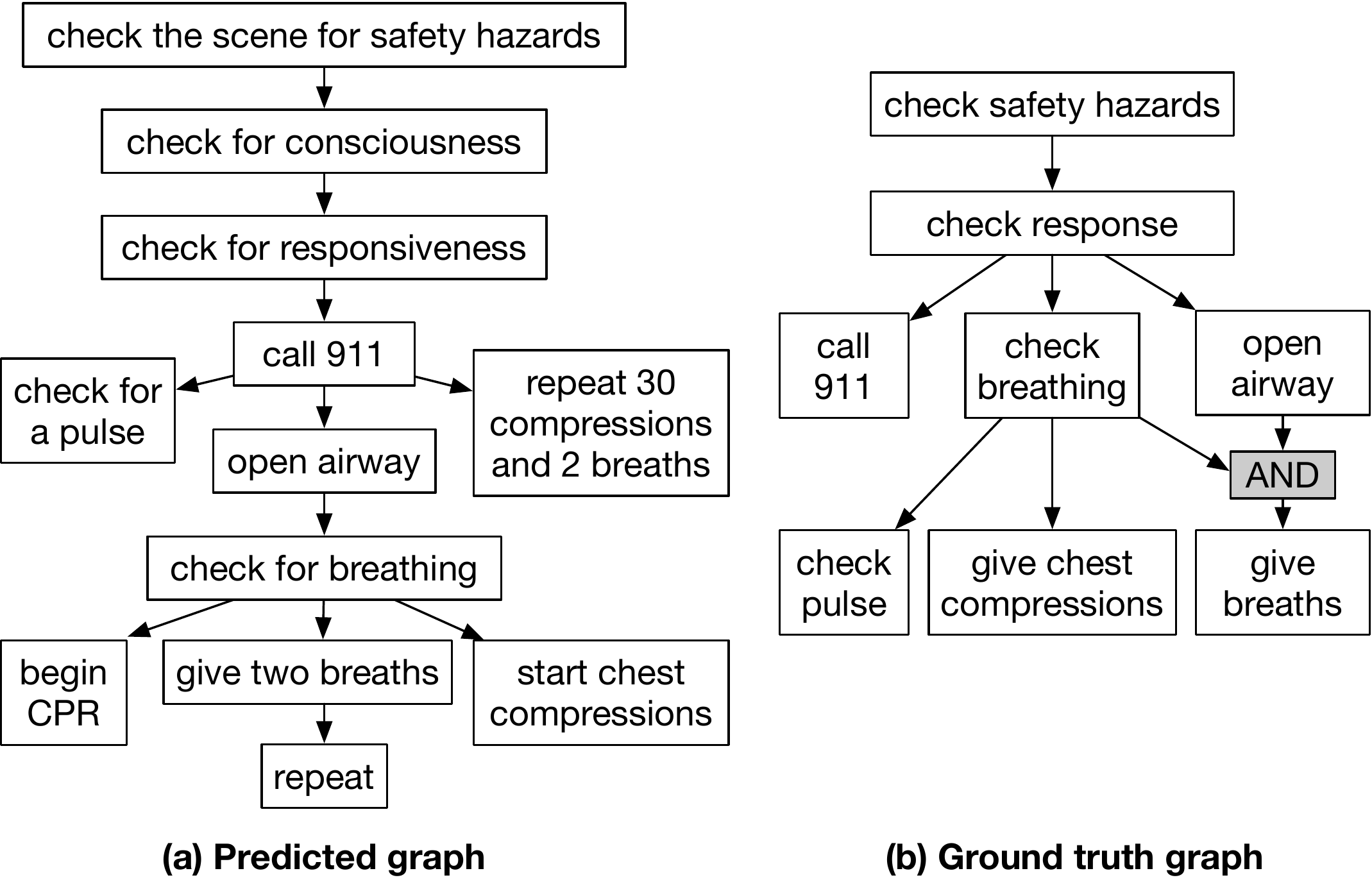}
        \caption{
Predicted (a) and ground truth (b) graphs for the \emph{perform cpr} task.
Edges indicate precondition relationships. %
Steps with multiple preconditions are represented using an \andorgraph{AND} node.
}
        \label{fig:cpr}
\end{figure}

There is a long history of work on reasoning about tasks, events and temporal ordering, broadly referred to as `script understanding' \citep{chambers2008unsupervised,regneri2010learning,modi2014inducing,schank2013scripts,pichotta2016learning}.
Recent formulations of script understanding problems include generating a sequence of steps from a given task description (e.g., bake a cake) \citep{lyu-etal-2021-goal,sancheti2021large,sun2022incorporating} and generating flow graphs from goal and event descriptions \citep{pal2021constructing,sakaguchi-emnlpf21}.
Script generation also manifests in interactive settings such as simulated embodied environments where agents are expected to reason about subgoals in order to complete tasks \cite{logeswaran2022few,huang2022language}.
Many of these prior approaches either fine-tune language models on human-annotated scripts or rely on knowledge encoded in language models to generate scripts.
In contrast, we attempt to use pre-trained language models as an information extraction system to perform zero-shot script inference from noisy ASR (Automatic Speech Recognition) transcriptions of instructional videos describing a task.
\vspace{-1em}
\begin{figure*}[!ht]
\small
\centering
\includegraphics[width=0.98\textwidth]{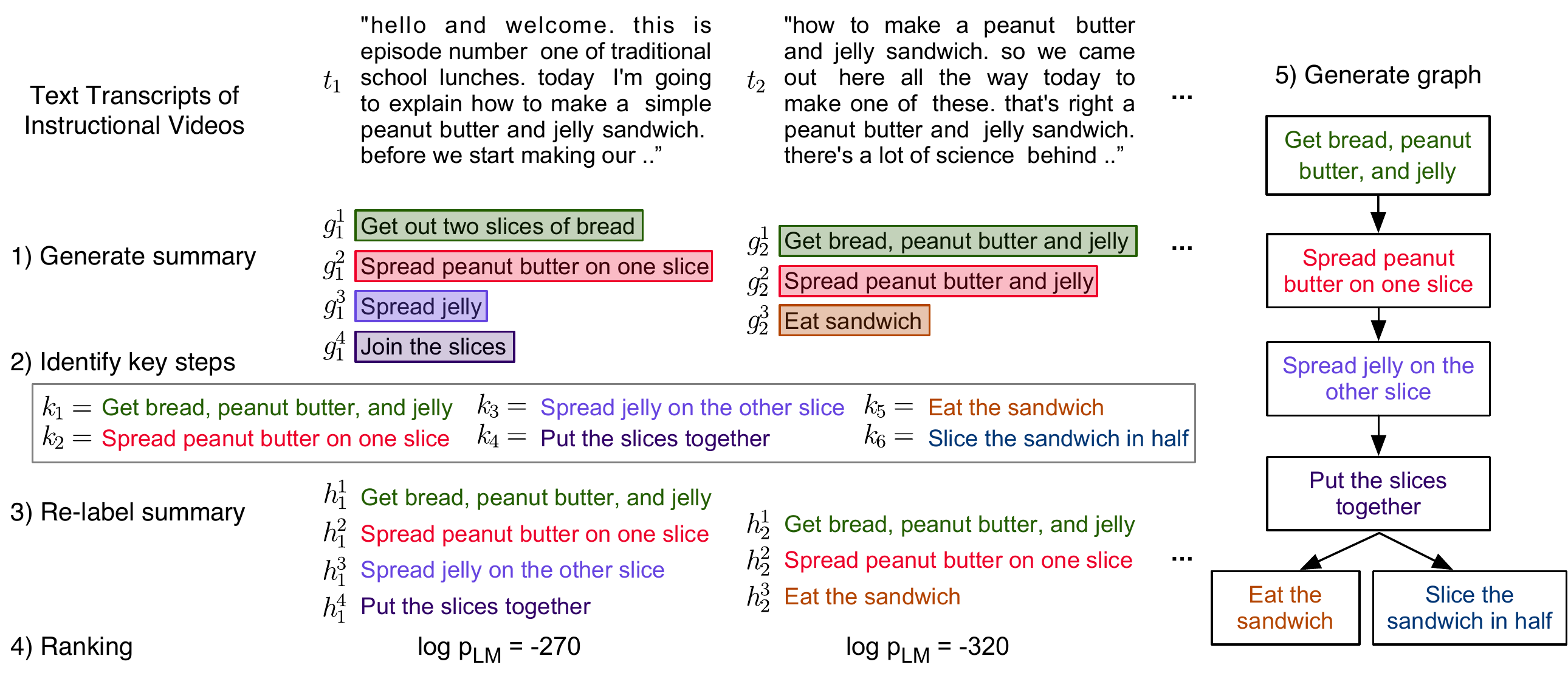}
\caption{
\textbf{Overview of our \Script~Generation Pipeline.} Given multiple text transcripts of a task, we 1) Summarize the steps described in the transcript, 2) Identify the key steps, 3) Re-label summary steps with key steps, 4) Rank key step sequences using a language model and 5) Consolidate top-k sequences to generate a \script~for the given task.  
}
\label{fig:overview}
\cutfigurebelow
\vspace{-0.5em}
\end{figure*}

Our focus in this work is to generate a directed graph that represents dependency relationships between the key steps relevant to a real-world task.
\Cref{fig:cpr} (a) shows a graph predicted by our approach for \emph{performing CPR}. 
An example dependency that can be read from the graph is that checking for safety hazards has to have happened before any other step (i.e., it is a \emph{precondition} that needs to be satisfied).
In this paper, we will use the term \emph{\script} to refer to such dependency graphs.

More formally, consider a real-world \emph{task} $\tau$.
We assume that multiple text \emph{transcripts} $t_1, \ldots, t_n$ describing how this task is performed are available.\footnote{Each transcript is a text document derived from an instructional video using Automatic Speech Recognition.} 
We assume that having access to such multiple transcripts helps robustly identify the dependencies between key steps so that an accurate \script~can be generated.
For instance, if step $y$ frequently follows step $x$, it is highly likely that step $x$ needs to happen before step $y$ (i.e., is a \textit{precondition}).
Our goal is to generate a \emph{\script}~for the given task $\tau$ which models these dependencies.
In particular, this involves (i) Identifying the \emph{key steps} $K = \{k_1, \ldots, k_m\}$ relevant to performing the task and (ii) Generating a graph with nodes $k_i$ and edges representing precondition relationships.%

Our contributions in this work are as follows.
\begin{enumerate} [topsep=0pt,itemsep=-1ex,partopsep=1ex,parsep=1ex,leftmargin=1em]
\item[\textbullet] We propose an unsupervised \script~generation pipeline that uses pretrained language models to infer key steps and their dependencies from multiple text descriptions of a real-world activity.
\item[\textbullet] We propose ranking and filtering mechanisms to improve the quality of generated \scripts.
\item[\textbullet] We demonstrate the effectiveness of the proposed approach compared to strong supervised and unsupervised baselines on two datasets. %
\end{enumerate}

\cutsectionabove
\section{Approach}
\cutsectionbelow

Our approach to \script~generation consists of multiple steps, illustrated in \Cref{fig:overview}.
First, we use an instruction-tuned language model to generate a summary of steps (in free-form text) from a transcript (\Cref{sec:summary}).
Given these \emph{summary step sequences} generated from multiple such transcripts for the task, we identify the key steps relevant to the task using a clustering approach (\Cref{sec:key steps}). 
We then re-label \emph{summary step sequences} using the identified key steps to obtain \emph{key step sequences} (\Cref{sec:labelsteps}) and rank them using a language model (\Cref{sec:ranking}).
Finally, we generate a \script~from the key step sequences (\Cref{sec:scriptgen}).

\cutsubsectionabove
\subsection{Generating Summary Steps}
\cutsubsectionbelow
\label{sec:summary}
The first step of our pipeline extracts a summary of steps $g_i = (g_i^1, g_i^2, \ldots)$ for performing the task described in each transcript $t_i$. %
We use an instruction-tuned language model for this purpose.
We prompt the model with a transcript, followed by a query such as \textit{`Based on this description list down the key steps for making coffee using short phrases.'} and let the model generate a completion.
We use the `Davinci' version of the InstructGPT \citep{ouyang-arxiv22} model in our experiments.
We observed that the model consistently generates the steps in the format `{\small 1. {\tt <step 1>}\textbackslash n 2. {\tt <step 2>}\textbackslash n ..}', occasionally using bullet points instead of numbers. 
The sentences $g_i^j$ on each line are extracted and treated as the summary steps identified from the transcript.
\Cref{appsec:summarysteps} shows example summary step sequences generated by InstructGPT.

\cutsubsectionabove
\subsection{Identifying Key Steps Relevant to the Task}
\cutsubsectionbelow
\label{sec:key steps}
Given summary step sequences $g_1, \ldots, g_n$ generated in the previous step, we seek to identify correspondences between steps in different summaries and capture the salient steps that appear frequently. %
We use a clustering approach for this purpose.
Sentences $g_i^j$ are represented as embeddings using a sentence encoder (We use the MiniLMv2 encoder from the SentenceTransformers library \citep{reimers-2019-sentence-bert,wolf2019huggingface}, which was identified as the best sentence embedding method for semantic search/retrieval). 
We obtain high-confidence clusters by identifying \textit{max cliques} -- clusters of sentences that are similar (determined by a threshold - cosine similarity $\geq 0.9$) to each other, and retain cliques with more than 5 sentences.
We noticed that this often yields multiple clusters that represent the same key step. 
For instance, the steps `\emph{fill the moka pot with water}' and `\emph{fill the bottom chamber with water}' represent the same key step of filling water, but are placed in different clusters. 
Identifying such redundant clusters based on sentence similarity alone is difficult.
We define the notion of \textit{sequence overlap} between two clusters -- how often a sentence from one cluster and a sentence from the other cluster appear in the same summary step sequence.
Intuitively, if two clusters have high inter-cluster similarity and low \textit{sequence overlap}, it is likely that they represent the same key step, and we merge the clusters. %
The resulting clusters obtained are treated as the key steps $k_1, \ldots, k_m$.\footnote{We use \textit{cluster} and \textit{key step} interchangeably. For visualization purposes we represent a cluster using a random sentence in the cluster.}
\Cref{appsec:keysteps} shows example clusters discovered for different tasks.

\cutsubsectionabove
\subsection{Re-labeling Summary Step Sequences}
\cutsubsectionbelow
\label{sec:labelsteps}
We re-label each summary step sequence $g$ (subscript $i$ dropped for brevity) with the identified key steps $k_1,\ldots,k_m$ to produce a \emph{key step sequence} $h$ using the greedy algorithm described in \Cref{alg:matching}.
The algorithm sequentially picks the most similar\footnote{defined as the maximum cosine similarity between $g^a$ and any sentence in cluster $k_b$ i.e., $\max_{s\in k_b} cos(g^a, s)$} candidate summary step and cluster pair $(g^a, k_b)$ at each step, assuming each key step only appears once in the sequence.
The process terminates when the highest cosine similarity drops below zero.

\cutsubsectionabove
\subsection{Ranking}
\cutsubsectionbelow
\label{sec:ranking}
One shortcoming of the labeling algorithm described in the previous section is that it does not take the sequential nature of steps into account.
To alleviate this issue, we use a language model to identify and filter the most promising key step sequences.
Specifically, we use $\text{log } p_\text{LM} (h|\text{prompt})$ as a measure of the quality of the key step sequence $h$, where we compute the likelihood of $h$ given a prompt similar to the prompt in \Cref{sec:summary} under a pre-trained language model. %
Multiple labelings are ranked based on this measure using a GPT2-XL model \citep{radford2019language} and the top-k are chosen as confident predictions for graph generation ($k = 75$\% in our experiments).\footnote{We use an open source language model considering API costs. Further, GPT2-XL led to decent ranking performance.}

\begin{algorithm}[t!]
\small
\caption{Key Step Sequence Inference}
\label{alg:matching}
\begin{algorithmic}
\Input{$g = (g^1, g^2, \ldots)$} \Comment{Summary step sequence}
\Input{$K = \{k_1, k_2, \ldots\}$} \Comment{Key steps}
\Statement{For each summary step identify most similar sentence from}
\Statement{each cluster:}
\Statement{$C_{ij} \gets \max_{s\in k_j} cos(g^i, s)$}
\Statement{$H_{ij} \gets \argmax_{s\in k_j} cos(g^i, s)$}
\Statement{$S \gets \{\}$} \Comment{Predicted alignments}
\While{$\max_{i,j} C_{ij} > 0$}
\State $a, b \gets \argmax_{i,j} C_{ij}$
\State $S \gets S \cup \{(a, b)\}$
\State $C_{aj} \gets 0$, $C_{ib} \gets 0 \quad \forall i, j$ 
\Statement{Sort $(a_i, b_i) \in S$ so that $a_1, a_2, \ldots$ are in increasing order}
\Statement{\textbf{Output} $h = (H_{a_1 b_1}, H_{a_2 b_2}, \ldots)$} \Comment{Key step sequence}
\end{algorithmic}
\end{algorithm}

\cutsubsectionabove
\subsection{\Script~Generation}
\cutsubsectionbelow
\label{sec:scriptgen}
We use an off-the-shelf algorithm \citep{sohn-iclr20, jang2022multimodal} which is based on Inductive Logic Programming (ILP) for constructing a graph from key step sequences $h_1,\ldots,h_n$.
Inuititively, the algorithm identifies a set of preconditions (which key step must precede another key step due to a causal relationship) most consistent with the key step sequences.
Details of the algorithm can be found in \Cref{appsec:scriptgen}.

\newcommand{\Arrow}[1]{%
\parbox{#1}{\tikz{\draw[->](0,0)--(#1,0);}}
}
\newcommand{\arr}{\Arrow{.5em}}
\newcommand{\colw}{1.4em}
\newcommand*\circled[1]{\tikz[baseline=(char.base)]{
            \node[shape=circle,draw,inner sep=0.5pt] (char) {#1};}}

\begin{table*}[!t]
\small
\centering
\begin{tabular}{l p{\colw} p{\colw} p{\colw} p{\colw} p{\colw} p{\colw} p{\colw} p{\colw} p{\colw} p{\colw} p{\colw} p{\colw} }
\toprule
\multicolumn{1}{c}{\multirow{3}{*}{Model}} 
& \multicolumn{6}{c}{\procel~(Accuracy $\uparrow$)} & \multicolumn{6}{c}{\crosstask~(Accuracy $\uparrow$)} \\
\cmidrule[0.8pt](lr){2-7} \cmidrule[0.8pt](lr){8-13}
& ~(a) & ~(b) & ~(c) & ~(d) & ~(e) & Avg & ~(f) & ~(g) & ~(h) & ~(i) & ~(j) & Avg \\
\midrule
\circled{1} Proscript \citep{sakaguchi-emnlpf21}
& 65.0      & 51.8      & 46.9      & 52.1      & 53.6      & 53.9      & 52.3      & \bf{89.6} & 57.0      & \bf{62.5} & 61.4      & 64.6      \\
\circled{2} ASR \arr Labels \arr Graph
& 52.5      & 57.1      & 78.1      & \bf{59.4} & 53.6      & 60.1      & 54.5      & 72.9      & \bf{71.1} & 56.2      & 54.5      & 61.8      \\
\circled{3} ASR \arr VPs \arr Labels \arr Graph
& 52.5      & 53.6      & 56.2      & \bf{59.4} & 53.6      & 55.1      & 54.5      & 72.9      & \bf{71.1} & 56.2      & 59.1      & 62.8      \\
\circled{4} ASR \arr GPT \arr Labels \arr Graph
& 68.8      & 69.6      & 87.5      & 53.1      & 55.4      & 66.9      & \bf{75.0} & 72.9      & 61.7      & \bf{62.5} & 65.9      & 67.6      \\
\circled{5} ASR \arr GPT \arr Labels \arr Rank \arr Graph
& \bf{76.2} & \bf{80.4} & \bf{90.6} & 51.0      & \bf{62.5} & \bf{72.1} & 72.7      & 77.1      & 61.7      & \bf{62.5} & \bf{68.2} & \bf{68.4} \\
\midrule
\circled{6} Ground-truth labels \arr Graph
& 82.5      & 83.9      & 78.1      & 78.1      & 96.4      & 83.8      & 79.5      & 89.6      & 68.0      & 68.8      & 72.7      & 75.7      \\
\bottomrule
\end{tabular}
\vspace*{-0.05in}
\caption{Graph prediction accuracy on \procel~and \crosstask~datasets. The tasks are 
(a) make PBJ sandwich
(b) change iphone battery
(c) perform CPR
(d) set up chromecast
(e) tie tie
(f) change tire
(g) make latte
(h) make pancakes
(i) add oil to car
(j) grill steak.
Baselines \circled{2}, \circled{3}, \circled{4} differ in the inputs used for key step labeling (i.e. \Cref{alg:matching}) -- they respectively use the ASR sentences, verb phrases extracted from the ASR and summary steps generated by GPT (\Cref{sec:summary}). 
\circled{5} is our proposed approach which includes top-k filtering (\Cref{sec:labelsteps}).
\circled{6} shows graph generation performance with ground truth key step sequences.
}
\vspace*{-0.1in}
\label{table:subgoalinference}
\end{table*}

\cutsectionabove
\section{Experiments}
\cutsectionbelow

\paragraph{Data}
We use \procel~\citep{elhamifar-iccv19} and \crosstask~\citep{zhukov2019cross} datasets in our experiments.
We experiment with five tasks from each dataset (considering API costs).
Task in these two datasets have respectively 13 and 7 key steps on average.
We use 60 instances for each task, where each instance is an instructional video along with its text transcript.
The transcripts have 565 tokens on average. %

\paragraph{Setup}
The datasets come with key steps annotations (i.e., $K$) for each task and \emph{key step sequence} annotations for each transcript. 
Our approach is unsupervised and does not make use of these annotations.
However, for evaluation purposes, we consider two settings.
The first setting assumes ground truth $K$ and evaluates the performance of the full pipeline ignoring the clustering component (since key steps are known). %
In the second setting, we use $K$ inferred from \Cref{sec:key steps} and perform qualitative comparisons with ground truth graphs.
Note that we did not use \emph{key step sequence} annotations from the datasets in either setting.

\cutsubsectionabove
\subsection{Results}
\cutsubsectionbelow

\paragraph{Known Key Steps}
\Cref{table:subgoalinference} compares our approach with baselines on graph prediction accuracy.
We use ground truth human annotated graphs from \citet{jang2022multimodal} for evaluation.
Proscript \citep{sakaguchi-emnlpf21} is a language model fine-tuned on manually curated script data.
Given a task description and a set of key steps, Proscript generates a partial order of the key steps.
In addition, we consider several variations of our approach as baselines in \Cref{table:subgoalinference}.

First, we observe that our unsupervised approach performs better than the proScript baseline which was explicitly trained on script data.
Second, using GPT generated summaries for labeling (\circled{4}) performs better than directly labeling the ASR sentences (\circled{2}) or verb phrases (VPs) extracted from the ASR (\circled{3}).
This baseline is inspired by prior work \citep{alayrac2016unsupervised,shen-cvpr21} which extract verb phrases from transcripts and attempt to identify salient actions using filtering/alignment mechanisms.
These approaches are susceptible to noise in the text data and are further limited by the assumption about each step being represented by a short verb phrase (extracted using syntax templates).
In contrast, we exploit large language models in order to extract key phrases from the transcript.
Third, we observe that ranking and filtering key step sequences using a language model (\circled{5}) further improves performance, with a significant improvement for \procel.
Finally, our approach comes closest to graphs generated from human annotated key step sequences in the datasets (\circled{6}).\footnote{Performance for ground truth labels is lower than 100\% due to noise in the human annotations, which is particularly prominent in~\crosstask.}
\Cref{appsec:summary,appsec:ranking} further present ablations showing the impact of modeling choices in our pipeline.

\paragraph{Unknown Key Steps}
Next, we consider the full pipeline where key steps are identified automatically.
Since ground truth reference \scripts~are unavailable in this case we perform a qualitative comparison of graphs generated using our approach and the ground truth, human annotated graph.
\Cref{fig:cpr,fig:overview} show predicted graphs for the tasks \emph{perform cpr} and \emph{make pbj sandwich}, respectively.

We observe that the predicted graph for \emph{perform cpr} is more detailed and fine-grained than the ground truth graph and captures many of the ground truth precondition relationships.
On the other hand, the graph for \emph{make pbj sandwich} is less fine-grained compared to the ground truth (\Cref{appfig:pbj} of \Cref{appsec:graph}).
For instance, the ground truth annotations distinguish between \emph{putting jelly on the bread} and \emph{spreading jelly on the bread}, whereas our approach treats them as a single step.
In addition, spreading peanut butter and spreading jelly are independent of each other and have no sequential dependency.
However, the predicted graph fails to capture this and assumes that the former is a precondition for the latter.
\Cref{appsec:graph} shows more examples of predicted graphs.

\cutsectionabove
\section{Conclusion and Limitations}
\cutsectionbelow

This work presented an unsupervised approach to generate \scripts~from text transcripts of instructional videos.
Our framework exploits multiple text transcripts which describe a task in order to robustly identify dependencies between key steps.
We demonstrated the effectiveness of our approach compared to several baselines.
A limitation of our work is the GPT API cost associated with scaling to a large number of tasks, which can be addressed by the use of large open-source language models.
Our work can be further improved by better integration between different components such as summary generation and clustering components which inform each other, which we leave to future work.

\bibliography{anthology,custom}
\bibliographystyle{acl_natbib}

\appendix
\onecolumn
\section{\Script~Generation}
\label{appsec:scriptgen}
\begin{figure}[!htp]
    \centering
    \includegraphics[width=0.98\textwidth]{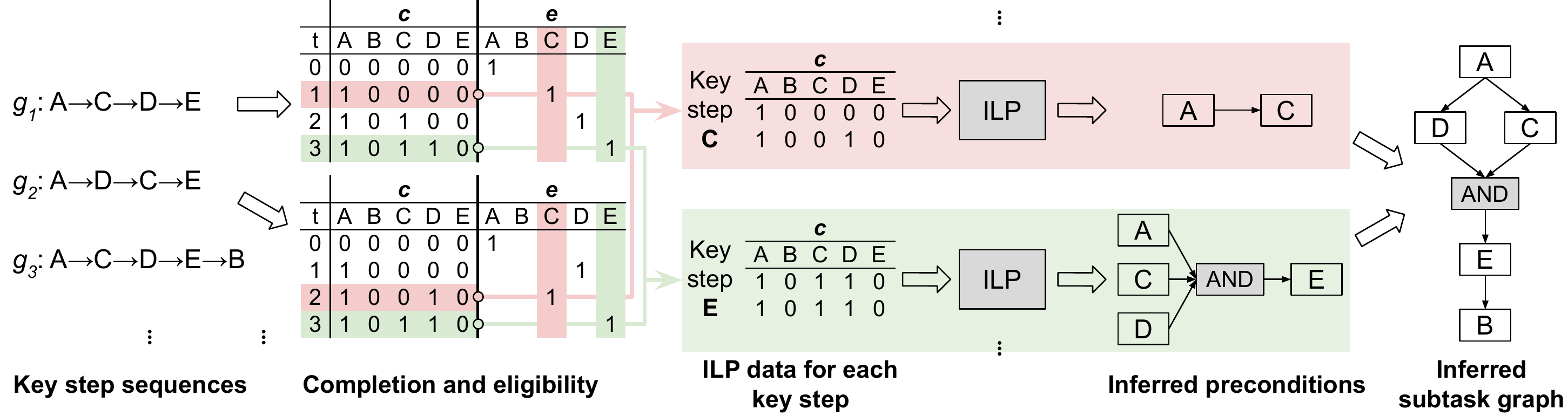}
    \caption{Procedure of predicting a \script~from key step sequences.}
    \label{fig:ilp-overview}
\end{figure}
We present details about the graph inference algorithm below.

\paragraph{Graph Representation}
Precondition describes the causal relationship between key steps relevant to a task and imposes a constraint on the order in which the key steps can be performed.
Formally, the precondition is defined as a logical expression that combines the key steps using \andorgraph{AND} and \andorgraph{OR} logic operations, which means \emph{all} or \emph{any} of certain key steps should be completed, respectively.  %
The precondition can be represented in disjunctive normal form (DNF) where multiple \andorgraph{AND} terms are combined with \andorgraph{OR} operations.
These preconditions can be compactly represented in the form of a graph.
The arguments of \andorgraph{AND} and \andorgraph{OR} operations in a precondition become the parents of corresponding \andorgraph{AND} and \andorgraph{OR} nodes in the graph, respectively.
For example, in \Cref{fig:cpr}, the precondition of step \subtask{give breaths} is \andorgraph{OR}(\andorgraph{AND}(\subtask{check breathing}, \subtask{open airway})).
Note that we omit \andorgraph{AND} and \andorgraph{OR} nodes with only one argument in the graph visualization for simplicity.

\paragraph{Graph Inference}
Given the set of known key steps $k_1, \ldots, k_m$ (\ie, vertices of the graph), graph inference aims to infer the preconditions (\ie, edges of the graph).
We first define the notions of \emph{completion} and \emph{eligibility} of key steps at a given point in time while the task is being performed.
We define the \emph{completion} vector $\smash{\mb{c}\in\{0, 1\}^{m}}$ as a binary vector where $c[p] \in \{0, 1\}; p\in\{1,\ldots,m\}$ represents whether key step $k_p$ was performed in the past.
Similarly we define the \emph{eligibility} vector $\smash{\mb{e}\in\{0, 1\}^{m}}$ as a binary vector where $e[p]$ represents whether key step $k_p$ is eligible to be performed (i.e., it's precondition is satisfied).
The completion and eligibility status of key steps will change over time as different key steps are performed to complete the task.
The precondition inference problem can be formulated as learning a function $\mb{e} = f_{\pcond}(\mb{c})$. %
In other words, precondition inference amounts to predicting the eligibility status of a key step given the completion status of all key steps.

Given a key step sequence $h = (h^1, h^2, \ldots)$, we convert it into a sequence of completion and eligibility vectors $((\mb{c}^1, \mb{e}^1), (\mb{c}^2, \mb{e}^2), \ldots)$ as described next.
We define the completion status $c^i[p]$ and eligibility status $e^i[p]$ of key step $k_p$ as follows.
If key step $k_p$ was completed in the past (i.e., there exists $j\le i$ s.t. $h^j\in k_p$), $c^i[p]$ is defined as 1 and 0 otherwise.
On the other hand, key step $k_p$ is considered eligible if $h^i\in k_p$ and it's eligibility is considered unknown otherwise.
Cases where eligibility is unknown are ignored by the algorithm.
See \Cref{fig:ilp-overview} for an illustration of this conversion process.

Given $\{(\mb{c}^j, \mb{e}^j)\}$ as training data, \citet{sohn-iclr20, jang2022multimodal} proposed an Inductive Logic Programming (ILP) algorithm which finds the graph $G$ that maximizes the data likelihood (\Cref{eq:ilp-objective})
\begin{align}
    \widehat{\pcond}
    &=\argmax_{\pcond} \prod_j{ p(\mb{e}^{j}|\mb{c}^{j}, G) }=\argmax_{\pcond} \sum_j{ \mathbb{I}[\mb{e}^{j} = f_{\pcond}(\mb{c}^{j})] }\label{eq:ilp-objective}
\end{align}
where $\mathbb{I}[\cdot]$ is the element-wise indicator function and $f_{\pcond}$ is the precondition function defined by the graph $G$, which predicts whether key steps are eligible from the key step completion vector $\mb{c}$.
The precondition function $f_G^p$ for key step $k_p$ (i.e., $e[p] = f_G^p(\mb{c})$) is modeled as a binary decision tree where each branching node chooses the best key step to predict whether the key step $k_p$ is eligible or not based on Gini impurity~\citep{breiman-routledge84}.
The precondition functions $f_G^1, \ldots,f_G^p$ learned for each key step $k_p$ induce a partial graph, which are consolidated to build the overall graph.
See \Cref{fig:ilp-overview} for an illustration of the process.

\paragraph{Graph Prediction Accuracy.}
\textit{Accuracy} (\Cref{eq:accuracy}) measures how often the output (\ie, eligibility) of the predicted and the ground-truth preconditions agree~\cite{sohn-iclr20}. $f_G^{p*}$ is the ground-truth precondition of the key step $k_p$. 
\begin{align}
    \text{Accuracy}=\frac{1}{m}\sum_{p=1}^{m}P\left(f_G^p(\mb{c})=f_G^{p*}(\mb{c}) 
    \right)
    \label{eq:accuracy}
\end{align}

\section{InstructGPT Summary Step Generation}
\label{appsec:summarysteps}

We present \textit{summary step sequences} generated by InstructGPT for the \emph{setup chromecast} task below, conditioned on text transcripts from the dataset.

{\small
\begin{enumerate}
\setlength{\itemsep}{-5pt}
    \item Go to Chromecast.com/setup
    \item Connect Chromecast to HDMI port
    \item Connect USB power cord to TV or power outlet
    \item Open Google Home App
    \item Follow on-screen instructions
\end{enumerate}
    
\begin{enumerate}
\setlength{\itemsep}{-5pt}
    \item Plug in the Chromecast to the TV.
    \item Connect the Chromecast to the Wi-Fi network.
    \item Use the Chromecast App to select what to cast.
\end{enumerate}
    
\begin{enumerate}
\setlength{\itemsep}{-5pt}
    \item Plug in the USB cable to the Chromecast.
    \item Connect the Chromecast to the HDMI port on the TV.
    \item Change the TV's input to the HDMI port that the Chromecast is connected to.
    \item Download the Chromecast App.
    \item Set up the Chromecast using the App.
    \item Choose the Wi-Fi network.
    \item Enter the Wi-Fi password.
    \item Cast from the computer by using the Chromecast extension in Google Chrome.
    \item Cast from the smartphone or tablet by using a compatible App.
\end{enumerate}
}

\noindent We present \textit{summary step sequences} generated by InstructGPT for the \emph{change iphone battery} task below, conditioned on text transcripts from the dataset.

{\small
\begin{enumerate}
\setlength{\itemsep}{-5pt}
\item  Turn off the phone
\item  Remove the bottom screws
\item  Lift up the screen
\item  Remove the metal plate
\item  Unclip the battery connector
\item  Pry up the battery
\item  Replace the battery
\item  Replace the metal plate
\item  Line up the screen
\item Snap the screen into place
\end{enumerate}

\begin{enumerate}
\setlength{\itemsep}{-5pt}
\item Unscrew the two pentalobe screws beside the Lightning jack.
\item Use a mini suction cup and place it right above the home button.
\item Use a guitar pick to gently rock back and forth until the screen starts lifting.
\item Unscrew the battery cover and remove the shield.
\item Unplug the existing battery by going under the metal flap with a flat edge.
\item Remove the adhesive that keeps the battery in place.
\item Place the new battery in the chassis and plug it in.
\item Place the battery cover back on and screw it in.
\item Lock the top edge of the screen in place.
\item Screw the bottom screws in place.
\end{enumerate}

\begin{enumerate}
\setlength{\itemsep}{-5pt}
\item Turn off the iPhone.
\item Remove the screws from the bottom of the phone.
\item Remove the screen from the phone.
\item Remove the battery connector.
\item Remove the adhesive strips from the old battery.
\item Attach the new adhesive strips to the new battery.
\item Place the new battery in the phone.
\item Reconnect the screen to the phone.
\item Replace the screws.
\item Turn on the phone.
\end{enumerate}
}

\newpage
\section{Key-steps identified}
\label{appsec:keysteps}

We show clusters/key steps identified by the clustering algorithm for the \emph{setup chromecast} task below.

\begin{enumerate}

{\small
\item 
\begin{itemize}
\setlength\itemsep{-2pt}
\item Connect the Chromecast to the Wi-Fi network
\item Connect to the same Wi-Fi network
\item Enter Wi-Fi password to connect Chromecast to Wi-Fi network
\item Join the Chromecast to the Wi-Fi network
\item Connect the Chromecast device to the Wi-Fi network
\item Connect the Chromecast to a Wi-Fi network
\item Connect to the Chromecast's Wi-Fi network
\item Connect the Chromecast to your Wi-Fi network
\end{itemize}

\item 
\begin{itemize}
\setlength\itemsep{-2pt}
\item Plug in the Chromecast to the TV
\item Plug in the Chromecast device to the TV
\item Connect the Chromecast to the TV
\end{itemize}

\item 
\begin{itemize}
\setlength\itemsep{-2pt}
\item Download the Google Home app
\item Download the Google Home application
\item Download the Google Home App
\end{itemize}

\item 
\begin{itemize}
\setlength\itemsep{-2pt}
\item Plug Chromecast into HDMI port and USB port on TV
\item Plug Chromecast into HDMI port on TV
\item Plug Chromecast into HDMI port and USB port for power
\item Plug Chromecast into the HDMI port on your TV
\item Plug Chromecast into power and HDMI port on TV
\item Plug in the Chromecast device to the HDMI port and USB port for power
\end{itemize}

\item 
\begin{itemize}
\setlength\itemsep{-2pt}
\item Go to Chromecast.com/setup
\item Go to chromecast.com/setup
\item Go to chromecast.com/setup on an Android device
\item Go to google.com/chromecast/setup
\item Go to google.com/chromecast/setup in Chrome browser
\end{itemize}

\item 
\begin{itemize}
\setlength\itemsep{-2pt}
\item Follow on-screen instructions to set up Chromecast
\item Follow the instructions on the app to set up Chromecast
\item Follow the prompts to set up the Chromecast
\item Follow the prompts to set up Chromecast
\end{itemize}

\item 
\begin{itemize}
\setlength\itemsep{-2pt}
\item Install the Chromecast App on your phone or tablet
\item Open the Google Home app on your phone or tablet
\item Install the Chromecast app on the phone
\item Install the Chromecast App on your Android device
\item Install the Chromecast App on a computer or mobile device
\end{itemize}

\item 
\begin{itemize}
\setlength\itemsep{-2pt}
\item Download Chromecast App
\item Download Chromecast app
\item Download the Chromecast App
\item Download the Chromecast app
\end{itemize}
}

\end{enumerate}

\newpage
\noindent We show clusters/key steps identified by the clustering algorithm for the \emph{change iphone battery} task below.

\begin{enumerate}

{\small
\item 
\begin{itemize}
\setlength\itemsep{-2pt}
\item remove the bottom two screws from the phone
\item Remove the screws at the bottom of the iphone
\item Remove the two pentalobe screws at the bottom of the phone
\item remove the two screws on the bottom of the iphone
\item Remove the two screws at the bottom of the iPhone
\end{itemize}

\item 
\begin{itemize}
\setlength\itemsep{-2pt}
\item remove battery
\item remove the battery
\item Remove battery
\item Remove the battery
\item Lift up the battery to remove it
\end{itemize}

\item 
\begin{itemize}
\setlength\itemsep{-2pt}
\item put in the new battery
\item Install the new battery
\item stick the new battery in
\item Insert the new battery
\item Put in new battery
\end{itemize}

\item 
\begin{itemize}
\setlength\itemsep{-2pt}
\item Pry up the frame of the screen with a pry tool
\item use a suction cup and sharp blade to pry open the screen case
\item use a suction cup and pry tool to remove the screen
\item use a pry tool to snap the latches and remove the screen
\item pry up very gently to separate the screen from the frame
\end{itemize}

\item 
\begin{itemize}
\setlength\itemsep{-2pt}
\item Turn off the phone
\item Turn off phone
\item Turn off the iPhone
\end{itemize}

\item 
\begin{itemize}
\setlength\itemsep{-2pt}
\item Remove the adhesive strips from the old battery
\item remove the adhesive from underneath the battery
\item use the fine tip curved tweezers to peel  up the edges of the two adhesive strips  at the bottom of the battery
\item remove the adhesive strips holding the battery in place
\end{itemize}

\item 
\begin{itemize}
\setlength\itemsep{-2pt}
\item Replace the screws
\item replace screws
\item Replace screws
\end{itemize}

\item 
\begin{itemize}
\setlength\itemsep{-2pt}
\item Lift up the screen with a suction cup
\item use the suction cup to pull the screen up gently
\item use a suction cup to pull up the screen
\item Use a suction cup to slightly lift the screen
\item Use a suction cup to apply upward pressure on the screen
\end{itemize}

\item 
\begin{itemize}
\setlength\itemsep{-2pt}
\item Remove the metal bracket and the two screws holding down the battery cable
\item remove the protective metal cover of the battery connector
\item Remove the two screws in the battery connector cover
\item remove the two screws on the shield that's covering the battery connector
\item unscrew the metal bracket holding the battery connector in place
\end{itemize}

\item 
\begin{itemize}
\setlength\itemsep{-2pt}
\item unscrew the four screws that cover the connectors for the screen
\item remove the cover plate that covers the screen connectors
\item Carefully dislodge the three connector tabs and set the screen aside
\item remove the metal cover and gently pry off the connectors of the screen one by one
\item Pull back the screen and remove the four screws securing the metal connector cover
\end{itemize}

}
\end{enumerate}

\newpage
\section{Generated graphs}
\label{appsec:graph}

We include generated graphs for other \procel~and \crosstask~tasks below.

\begin{figure}[h!]
     \centering
     \begin{subfigure}[c]{0.47\textwidth}
         \centering
         \includegraphics[width=\textwidth]{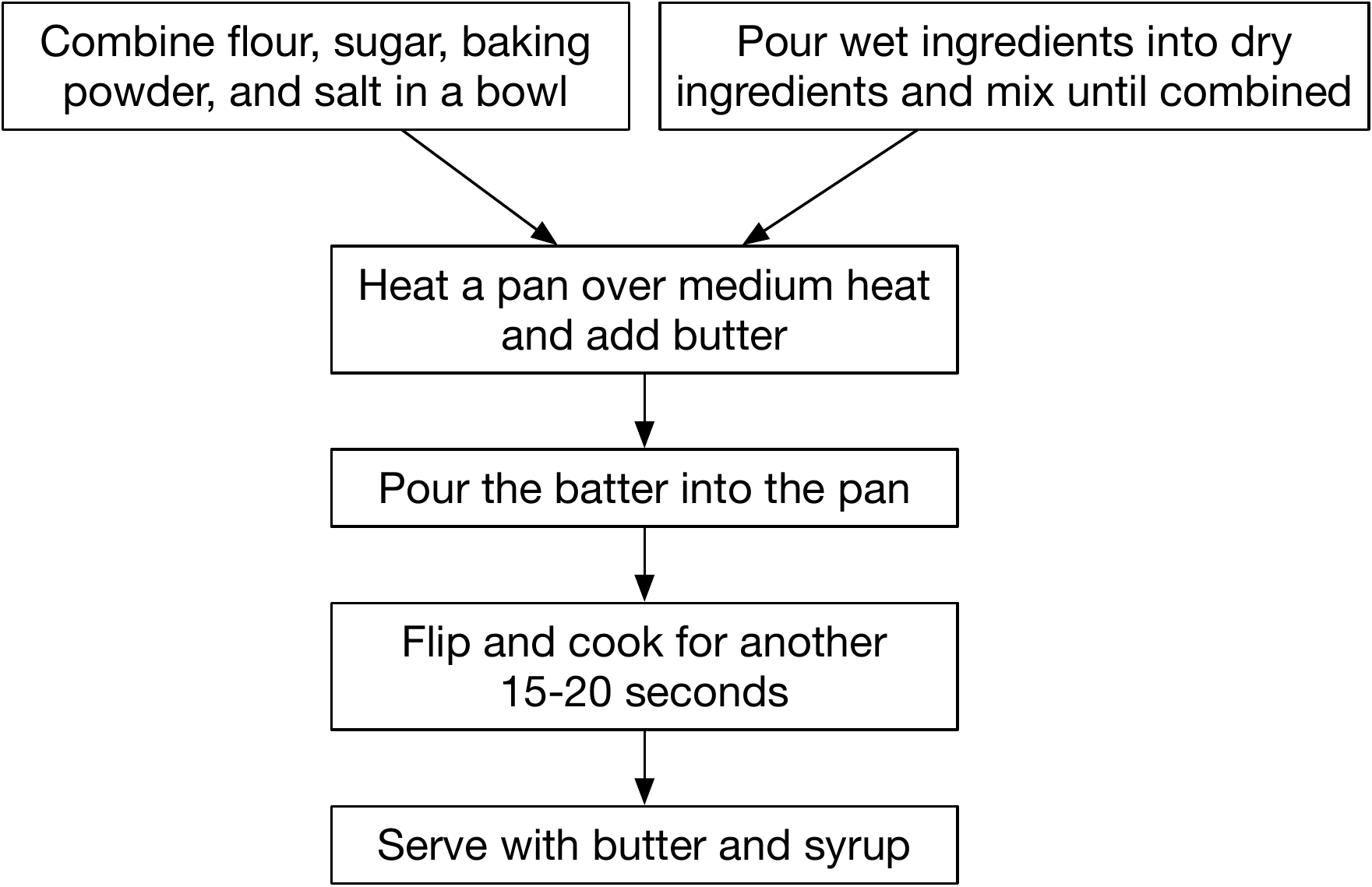}
         \caption{Predicted graph}
         \label{appfig:pancakespred}
     \end{subfigure}
     \hfill
     \begin{subfigure}[c]{0.50\textwidth}
         \centering
         \includegraphics[width=\textwidth]{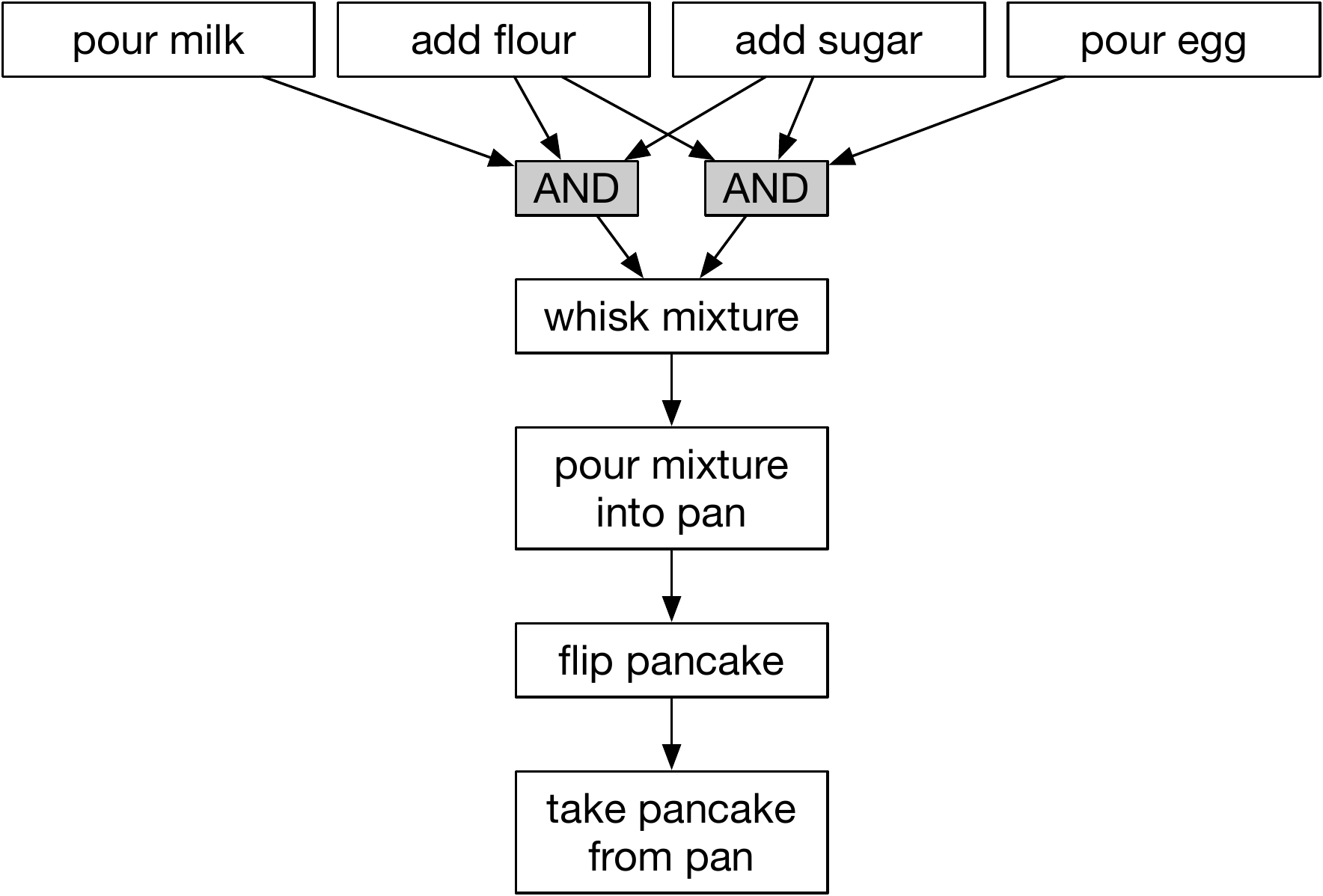}
         \caption{Ground truth grpah}
         \label{appfig:pancakesgt}
     \end{subfigure}
     \caption{Predicted (a) and ground truth (b) graphs for the \emph{make pancakes} task.}
\end{figure}

\begin{figure}[h!]
     \centering
     \begin{subfigure}[c]{0.50\textwidth}
         \centering
         \includegraphics[width=\textwidth]{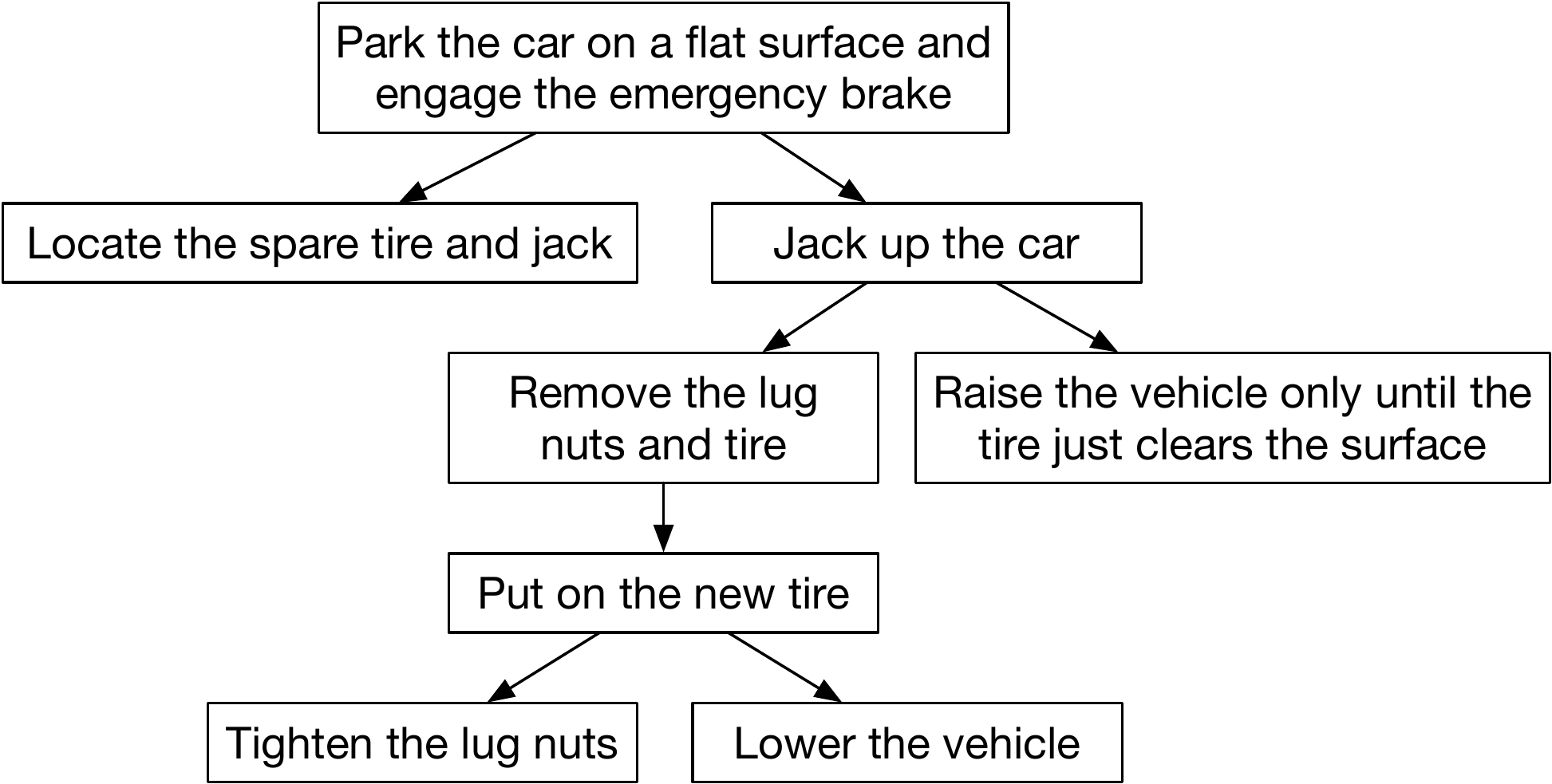}
         \caption{Predicted graph}
         \label{appfig:tirepred}
     \end{subfigure}
     \hfill
     \begin{subfigure}[c]{0.47\textwidth}
         \centering
         \includegraphics[width=\textwidth]{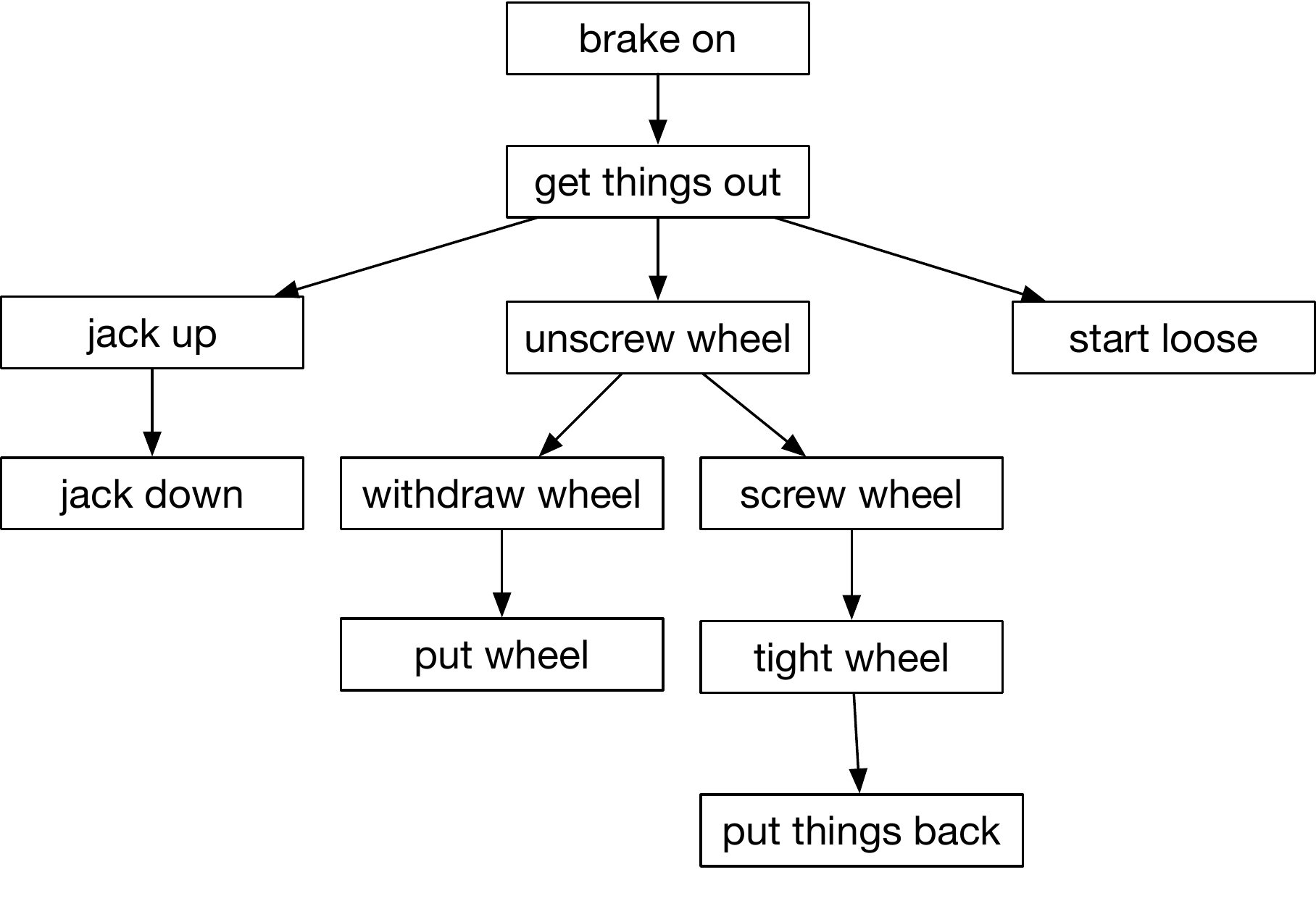}
         \caption{Ground truth grpah}
         \label{appfig:tiregt}
     \end{subfigure}
     \caption{Predicted (a) and ground truth (b) graphs for the \emph{change tire} task.}
\end{figure}

\begin{figure}[h!]
     \centering
     \begin{subfigure}[c]{0.45\textwidth}
         \centering
         \includegraphics[width=\textwidth]{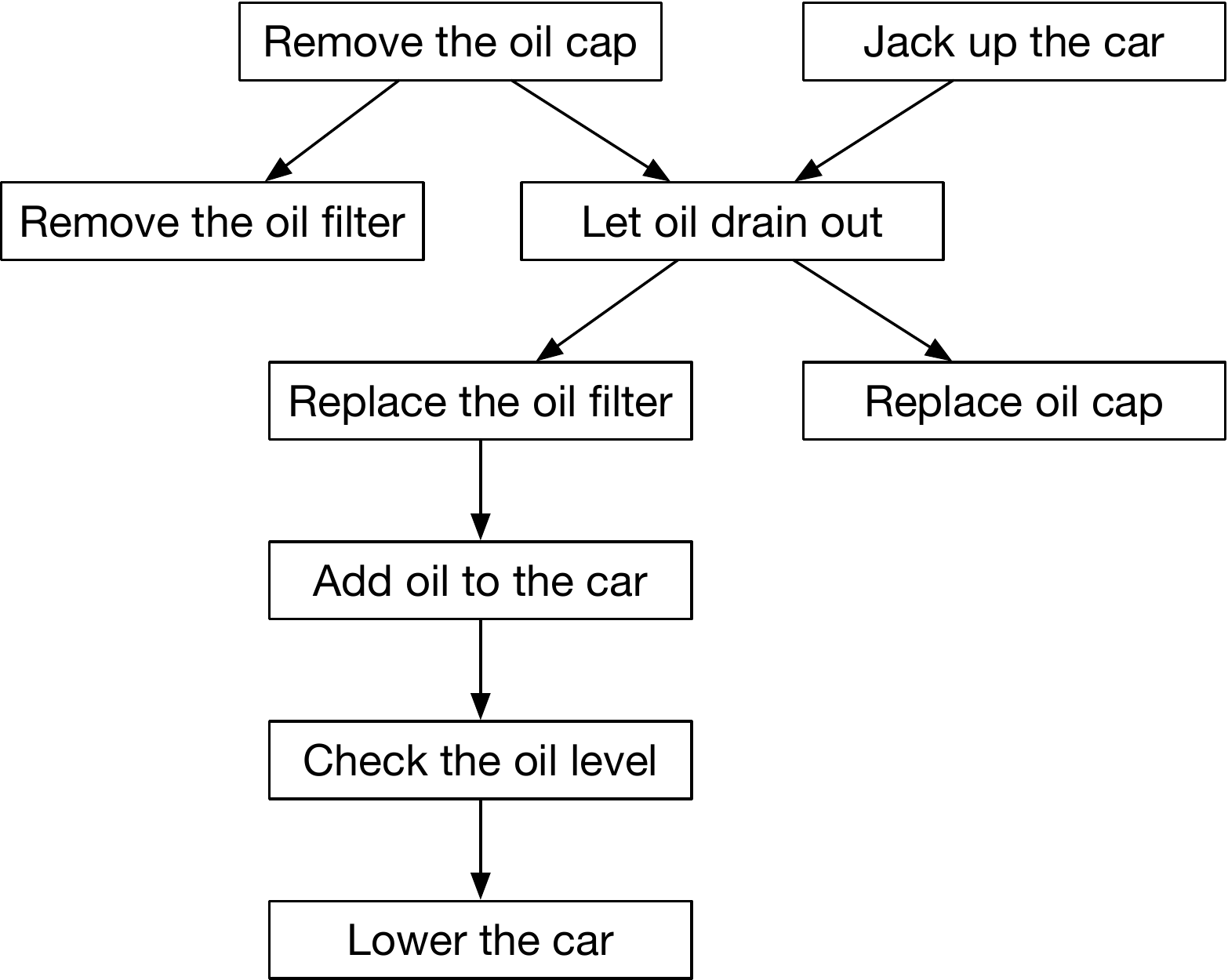}
         \caption{Predicted graph}
         \label{appfig:oilpred}
     \end{subfigure}
     \hspace{2em}
     \begin{subfigure}[c]{0.30\textwidth}
         \centering
         \includegraphics[width=\textwidth]{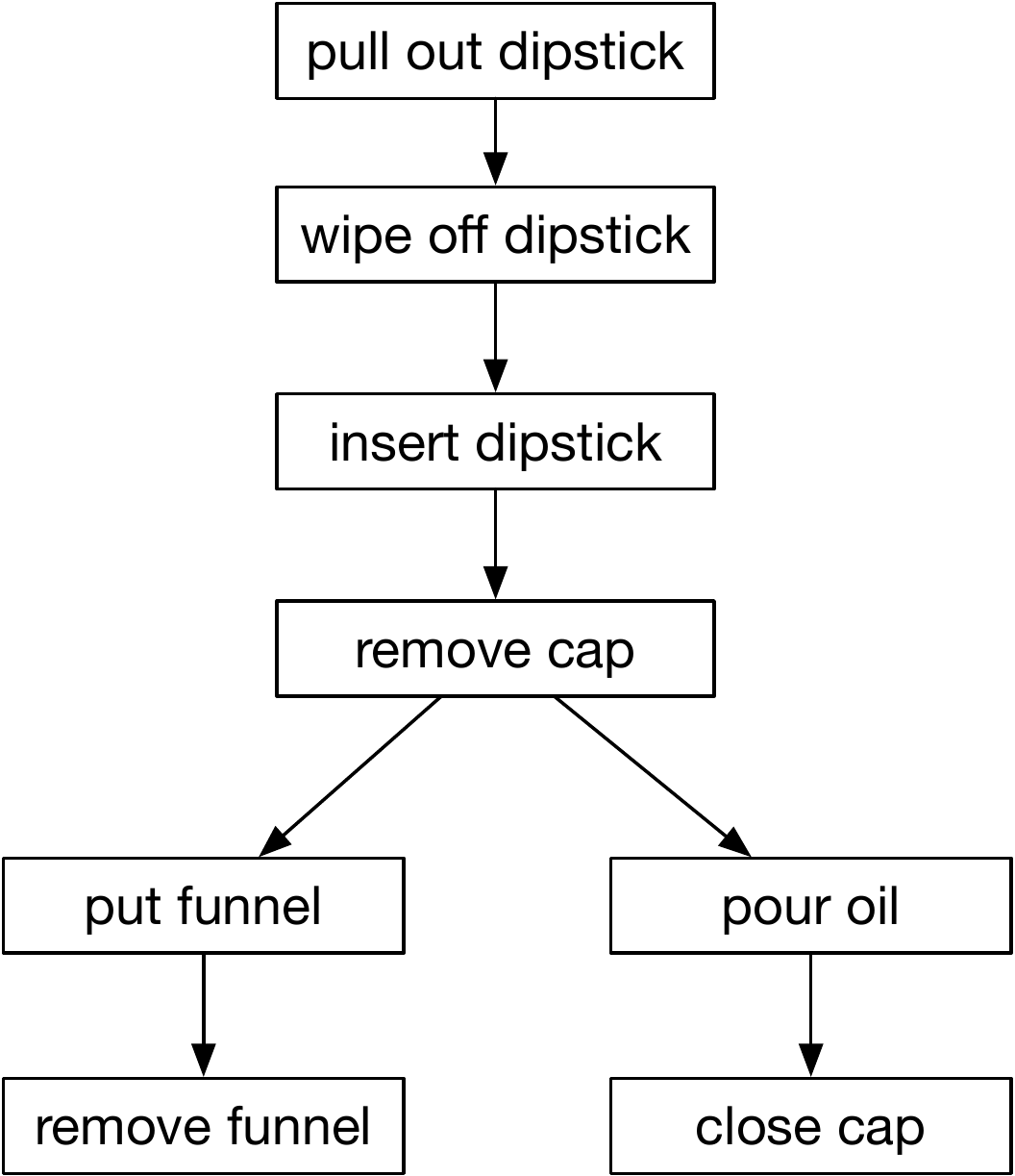}
         \caption{Ground truth grpah}
         \label{appfig:oilgt}
     \end{subfigure}
     \caption{Predicted (a) and ground truth (b) graphs for the \emph{add oil to your car} task.}
\end{figure}

\begin{figure}[h!]
     \centering
     \begin{subfigure}[c]{0.58\textwidth}
         \centering
         \includegraphics[width=\textwidth]{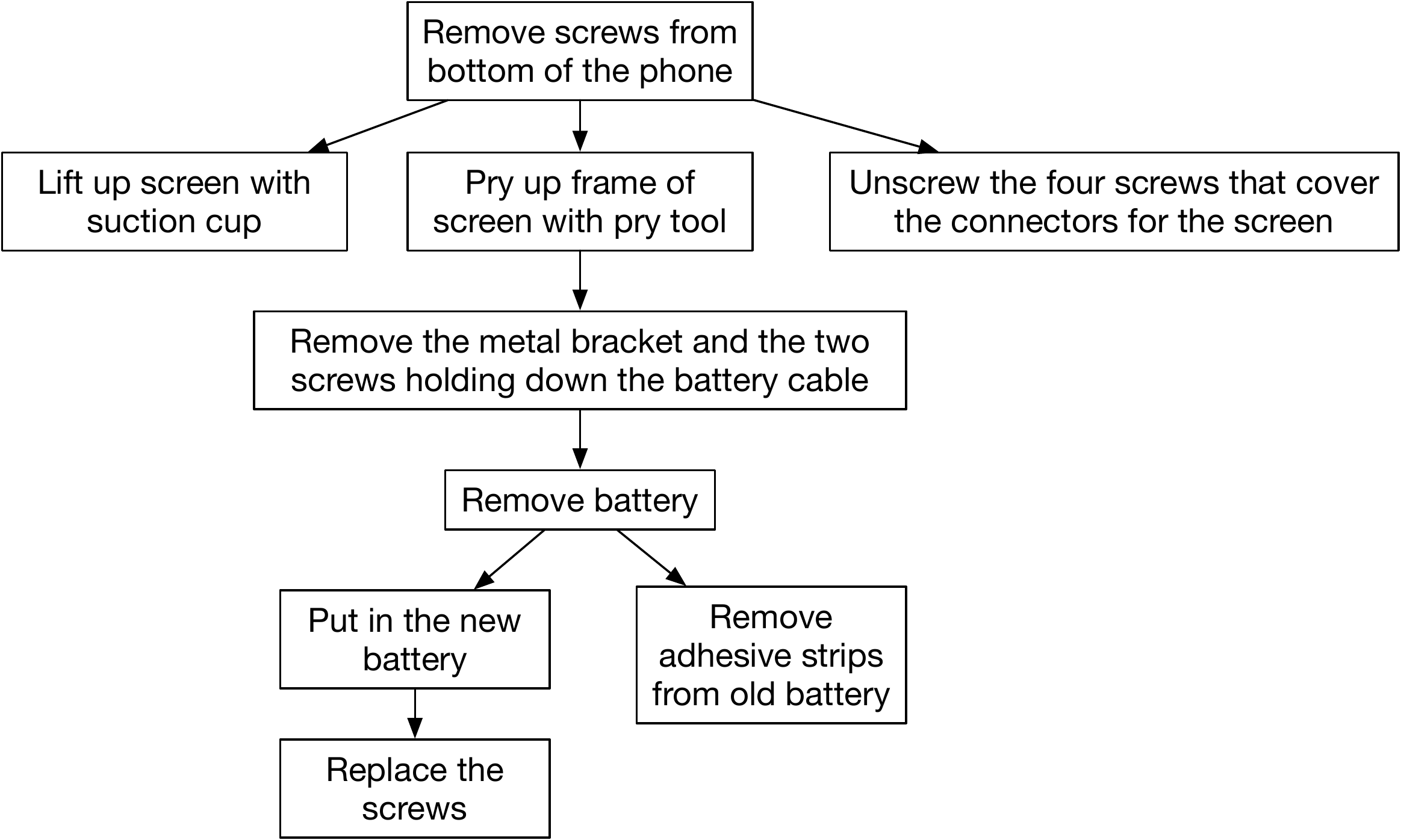}
         \caption{Predicted graph}
         \label{appfig:iphonepred}
     \end{subfigure}
     \hfill
     \begin{subfigure}[c]{0.40\textwidth}
         \centering
         \includegraphics[width=\textwidth]{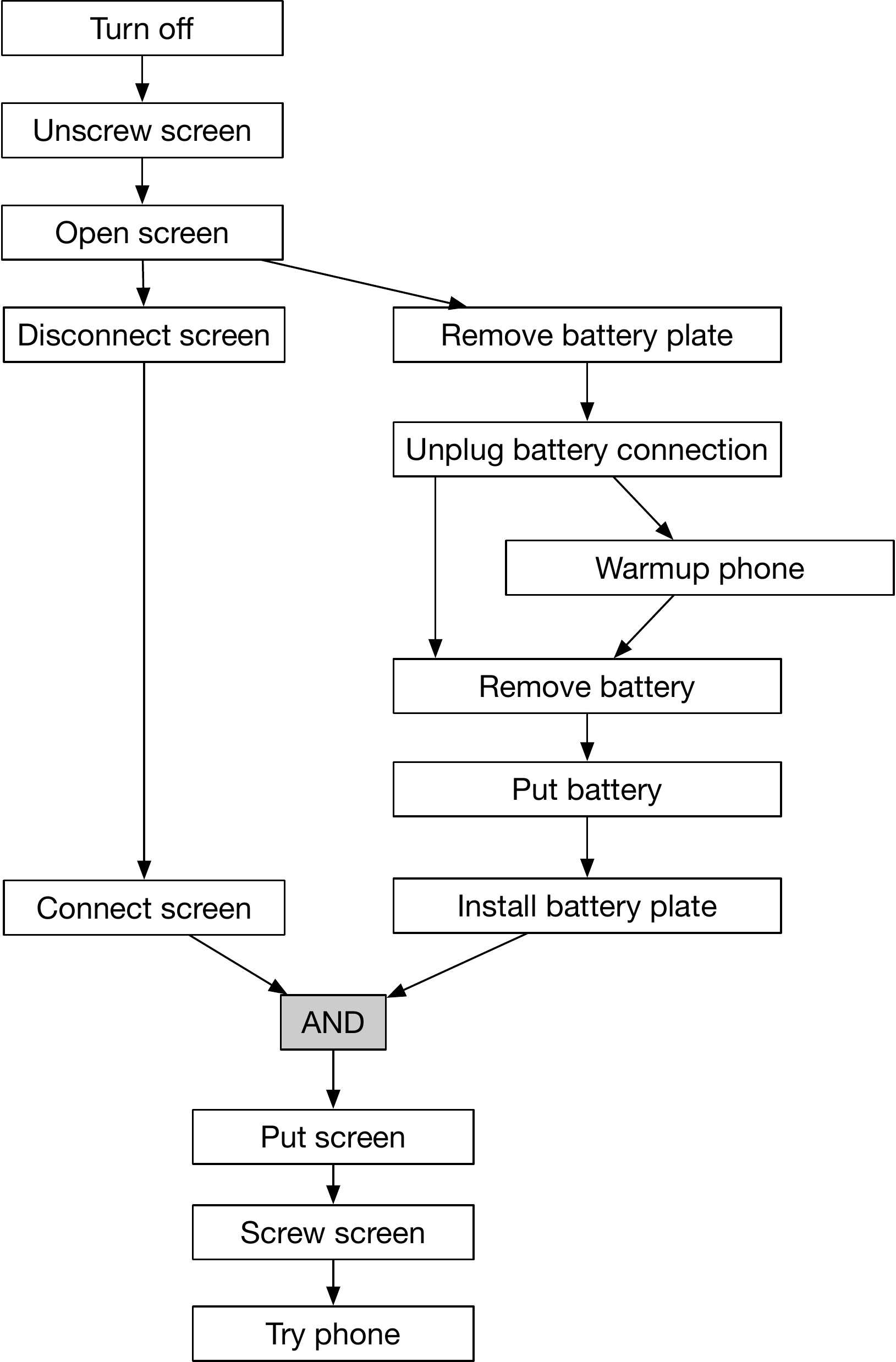}
         \caption{Ground truth grpah}
         \label{appfig:iphonegt}
     \end{subfigure}
     \caption{Predicted (a) and ground truth (b) graphs for the \emph{change iphone battery} task.}
\end{figure}

\begin{figure}[h!]
     \centering
     \begin{subfigure}[c]{0.43\textwidth}
         \centering
         \includegraphics[width=\textwidth]{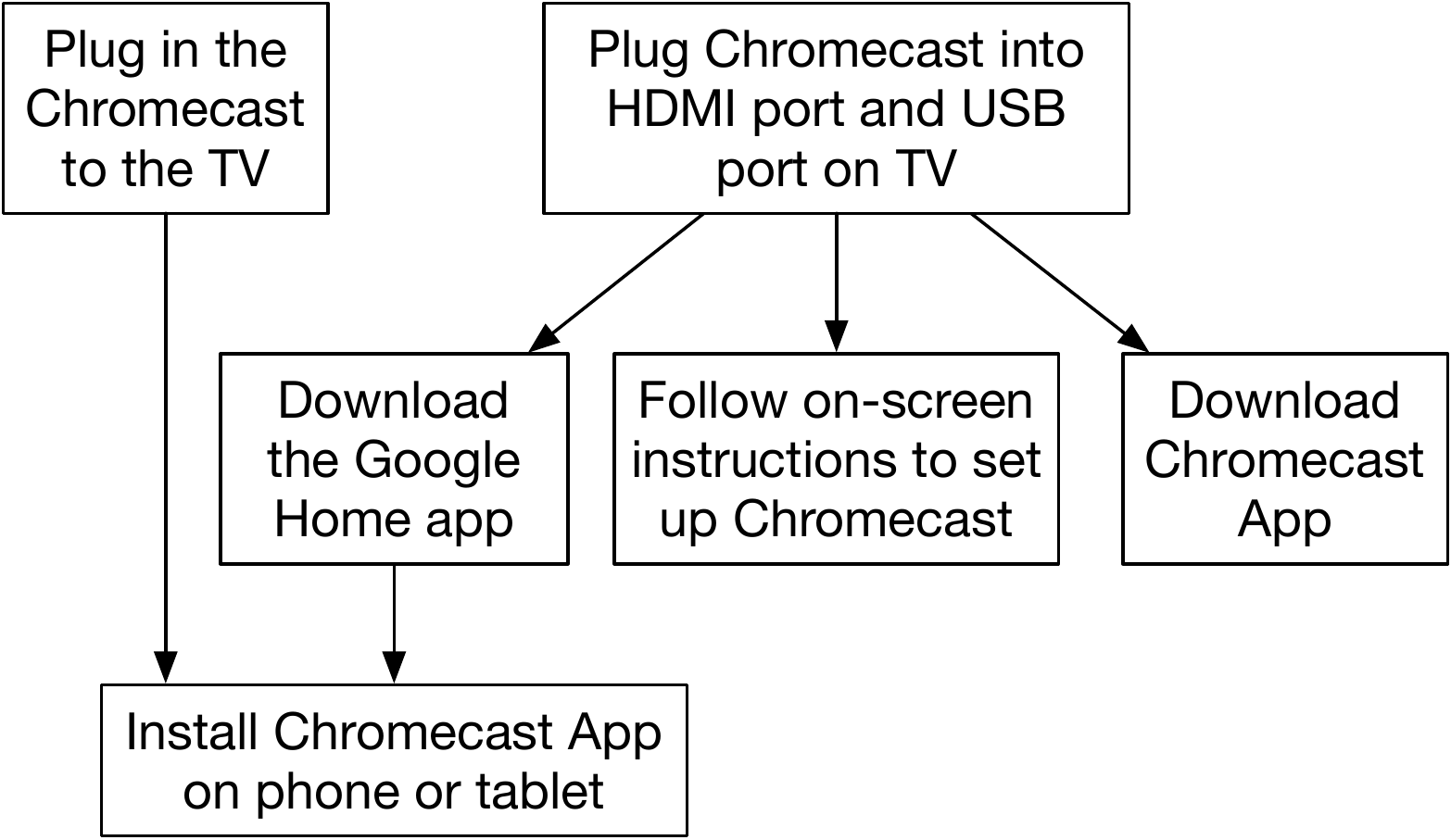}
         \caption{Predicted graph}
         \label{appfig:chromepred}
     \end{subfigure}
     \hfill
     \begin{subfigure}[c]{0.54\textwidth}
         \centering
         \includegraphics[width=\textwidth]{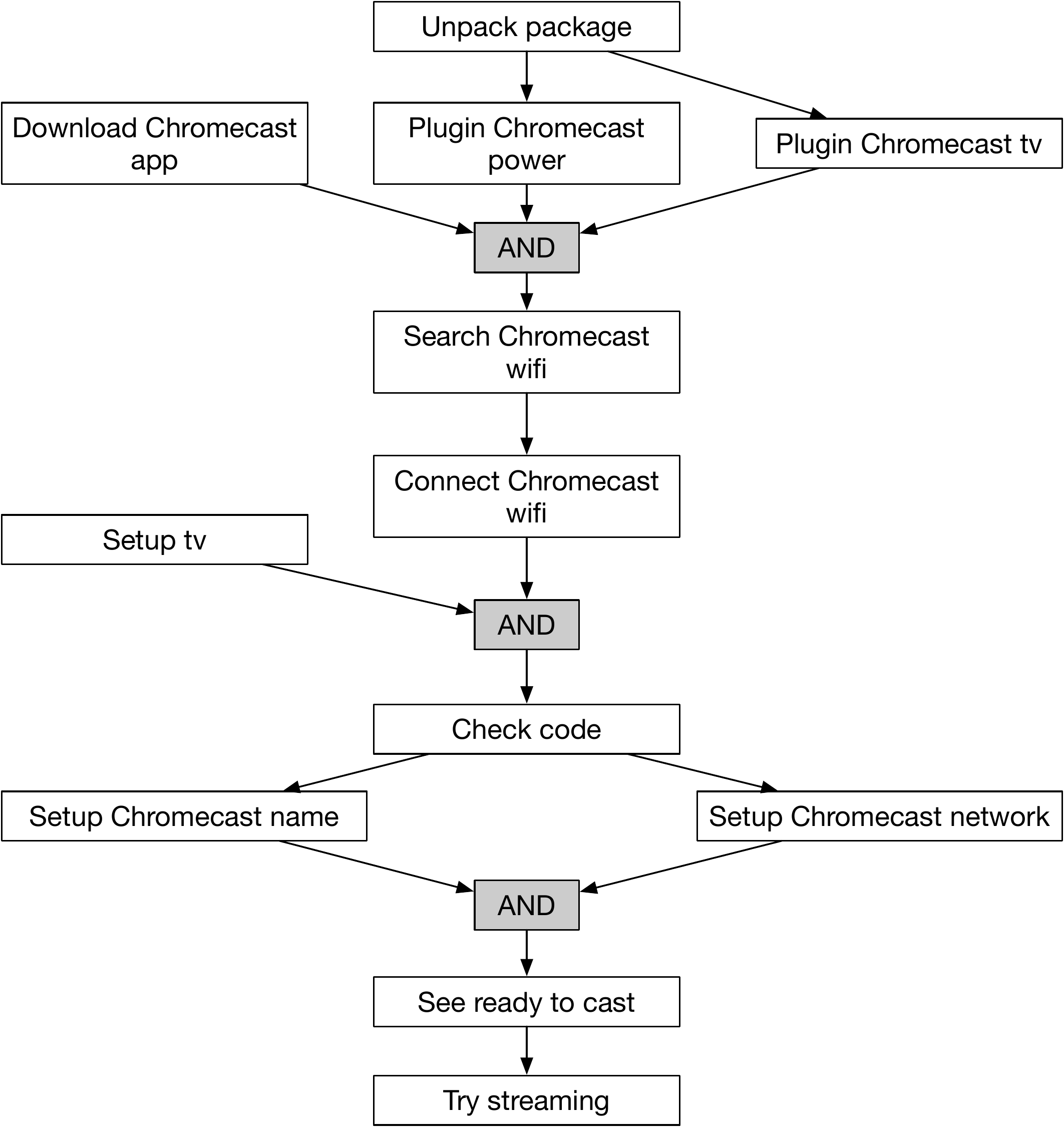}
         \caption{Ground truth grpah}
         \label{appfig:chromegt}
     \end{subfigure}
     \caption{Predicted (a) and ground truth (b) graphs for the \emph{setup chromecast} task.}
\end{figure}

\clearpage

\begin{figure}[h!]
     \centering
     \begin{subfigure}[c]{0.30\textwidth}
         \centering
         \includegraphics[width=\textwidth]{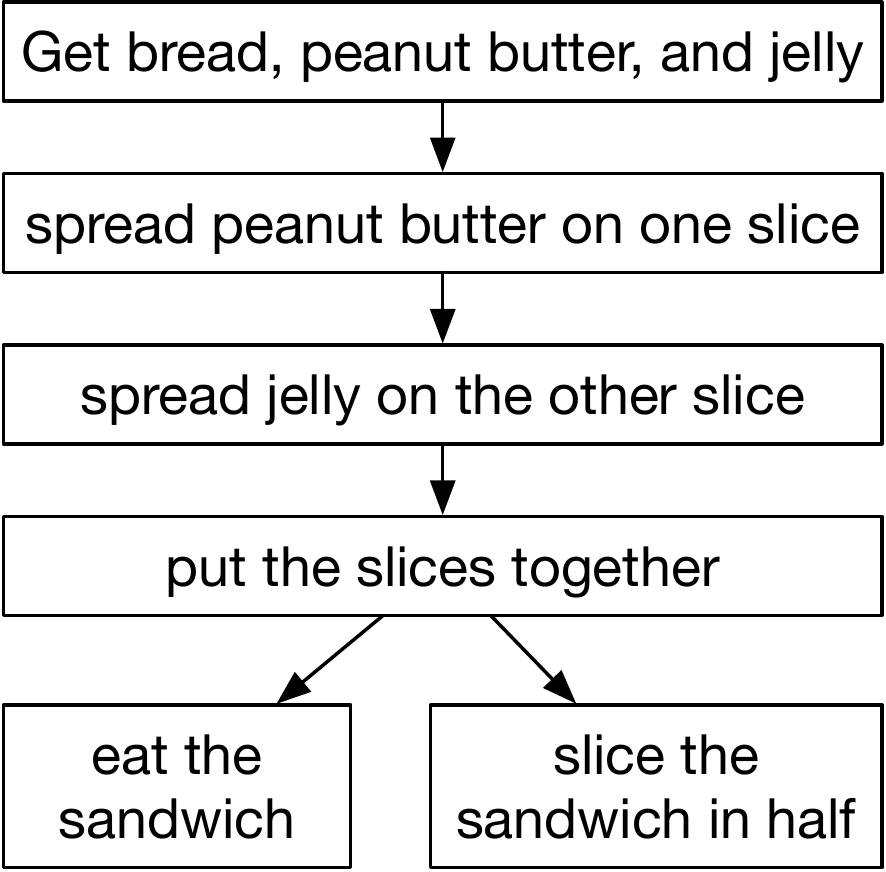}
         \caption{Predicted graph}
         \label{appfig:pbjpred}
     \end{subfigure}
     \hspace{5em}
     \begin{subfigure}[c]{0.40\textwidth}
         \centering
         \includegraphics[width=\textwidth]{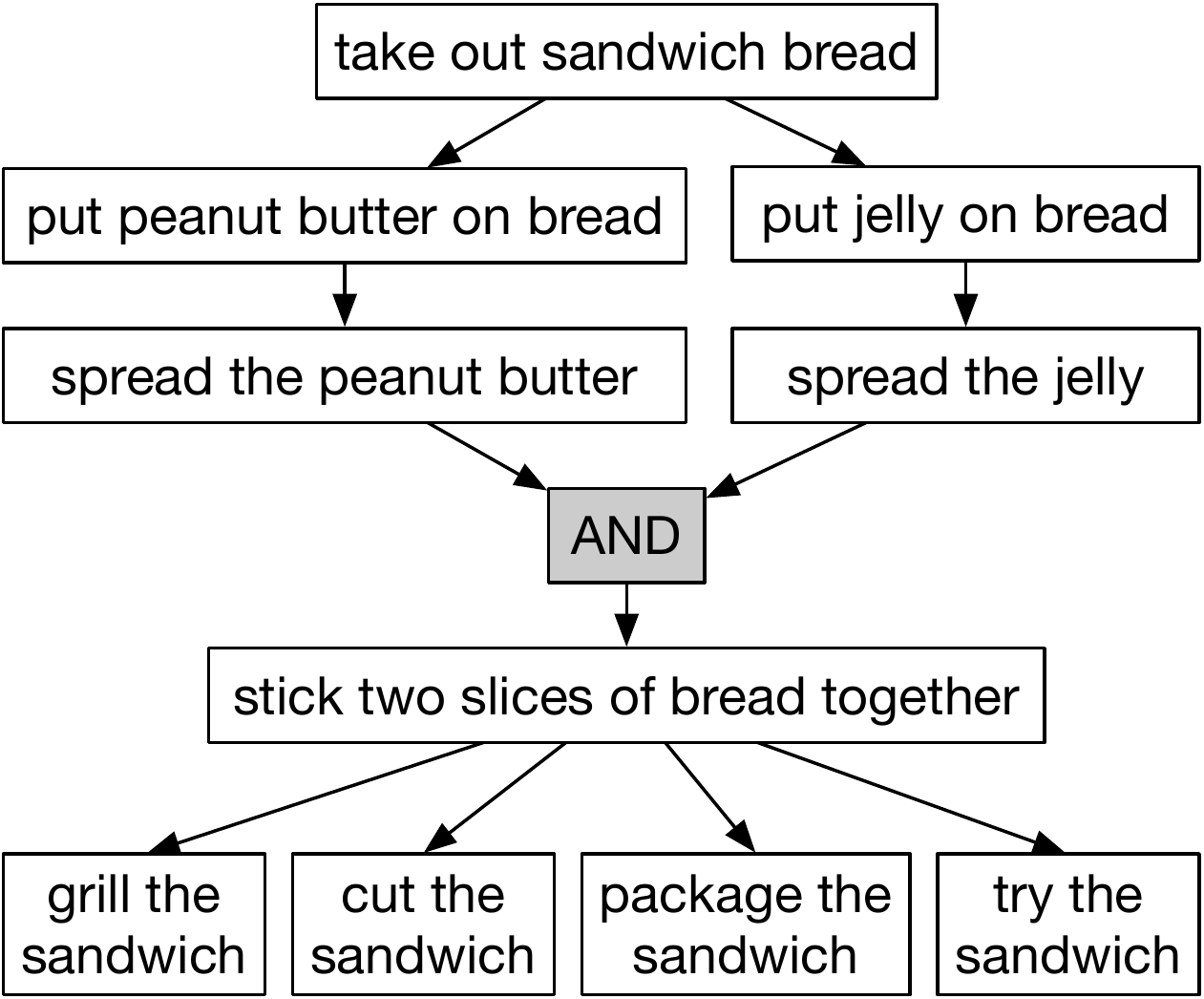}
         \caption{Ground truth grpah}
         \label{appfig:pbjgt}
     \end{subfigure}
     \caption{Predicted (a) and ground truth (b) graphs for the \emph{make PBJ sandwich} task.}
\label{appfig:pbj}
\end{figure}

\section{Choice of Summary Step Sequence Generation Model}
\label{appsec:summary}
We perform an ablation to study the effect of the model used to generate summary step sequences from transcripts.
We replace the InstructGPT model \citep{ouyang-arxiv22} with a FLAN-T5 model \citep{chung2022scaling} and evaluate graph prediction performance. 
We find that InstructGPT consistently outperforms FLAN-T5 across all the tasks. 
In addition, we found that plain language models (not fine-tuned with instructions) struggled to produce usable summaries.
This shows that models trained with instructions and human-preference data are better at producing \scripts~from transcripts compared to other forms of supervision such as language modeling and supervised multi-task training with NLP tasks.

\begin{table}[!h]
\centering
\begin{tabular}{l c c c c c c}
\toprule
Summary step generator & (a) & (b) & (c) & (d) & (e) & Avg \\
\midrule
FLAN-T5 \citep{chung2022scaling} & 60.0 & 64.3 & 87.5 & 49.0 & 57.1 & 63.6 \\
InstructGPT \citep{ouyang-arxiv22} & \textbf{76.2} & \textbf{80.4} & \textbf{90.6} & \textbf{51.0} & \textbf{62.5} & \textbf{71.1} \\
\bottomrule
\end{tabular}
\vspace*{-0.05in}
\caption{
Graph prediction accuracy on the Procel dataset when different models are used for summary step generation. The tasks are 
(a) make PBJ sandwich
(b) change iphone battery
(c) perform CPR
(d) set up chromecast
(e) tie tie.
}
\label{table:summary}
\end{table}

\section{Choice of Ranking Language Model}
\label{appsec:ranking}
We perform an ablation to study to understand the impact of the choice of language model for the ranking process in \Cref{sec:ranking}.
We present the average performance on tasks in the \procel~dataset with different language model choices in \Cref{table:ranking}.
First, we find that performance does not degrade much when switching to a smaller model in the GPT2 family.
Second, we notice that scale alone does not guarantee better ranking performance as the larger GPT-J model \citep{gpt-j} is inferior to the GPT2 models.
These findings suggest that the choice of pre-training data influences the script knowledge present in a model and can be more important than model scale.

\begin{table}[!h]
\centering
\begin{tabular}{l c c}
\toprule
Language Model & Parameter Count & Performance \\
\midrule
GPT2-Medium \citep{radford2019language} & 345M  & 70.96 \\
GPT2-XL \citep{radford2019language}     & 1.5B  & \textbf{72.14} \\
GPT-J \citep{gpt-j}                     & 6B    & 68.80 \\
\bottomrule
\end{tabular}
\vspace*{-0.05in}
\caption{
Ranking performance of different Language Models on the Procel dataset.
}
\label{table:ranking}
\end{table}

\end{document}